\documentclass[sigconf,natbib]{acmart}
\usepackage{multirow,multicol}
\usepackage{color,tabularx, colortbl,amsmath,bm,arydshln}
\usepackage{kantlipsum}
\usepackage{booktabs,url,tcolorbox,mathtools} 
\usepackage[]{graphicx}
\usepackage{subfigure}
\usepackage{caption}
\usepackage{scalerel}

\newcommand{\ttiSet}[1]{\mathcal{#1}}
\newcommand{\images}[0]{\ttiSet{X}}
\newcommand{\numgen}[0]{k}
\newcommand{\image}[0]{x}
\newcommand{\trajectory}[0]{\pi}
\newcommand{\relevance}[0]{f^*}
\newcommand{\rbp}[0]{\text{RBP}}
\newcommand{\err}[0]{\text{ERR}}

\newcommand{\novrbp}[0]{\text{RBP}_{\novelty}}
\newcommand{\noverr}[0]{\text{ERR}_{\novelty}}
\newcommand{\urbp}[0]{\novrbp^u}
\newcommand{\uerr}[0]{\noverr^u}

\newcommand{\patience}[0]{\gamma}
\newcommand{\satiation}[0]{g^*}
\newcommand{\salience}[0]{\sigma}
\newcommand{\novelty}[0]{\eta}
\newcommand{\targetimage}[0]{\hat{\image}}
\newcommand{\generatedimage}[0]{\image}
\newcommand{\prob}[0]{\text{Pr}}

\newcommand{\prompt}[0]{q}
\newcommand{\system}[0]{s}

\newcommand{\evalmetric}[0]{\mu}

\newcommand{\MSCOCO}[0]{\texttt{MS-COCO}}
\newcommand{\LNCOCO}[0]{\texttt{LN-COCO}}
\newcommand{\prompts}[0]{\texttt{Prompts}}

\newcommand{\promptutility}[0]{u_q}

\newcommand{\gridWidth}[0]{w}
\newcommand{\gridHeight}[0]{h}

\newcommand{\evalq}[0]{\mu(\system(\prompt),\promptutility)}

\newcommand\reallywidehat[1]{\arraycolsep=0pt\relax%
\begin{array}{c}
\stretchto{
  \scaleto{
    \scalerel*[\widthof{\ensuremath{#1}}]{\kern-.5pt\bigwedge\kern-.5pt}
    {\rule[-\textheight/2]{1ex}{\textheight}} 
  }{\textheight} %
}{0.5ex}\\           
#1\\                 
\rule{-1ex}{0ex}
\end{array}
}

\newcommand{\imageembedding}[0]{\phi}

\newcommand{\similarity}[2]{\left\langle #1,#2 \right\rangle}
\usepackage{relsize,blkarray}
\usepackage{algpseudocode}
\usepackage{amsmath}
\usepackage{verbatim}

\makeatletter

\newcommand\addauthornote[1]{%
  \if@ACM@anonymous\else
    \g@addto@macro\addresses{\@addauthornotemark{#1}}%
  \fi}

\newcommand\@addauthornotemark[1]{\let\@tmpcnta\c@footnote
   \setcounter{footnote}{#1}\addtocounter{footnote}{-1}
    \g@addto@macro\@currentauthors{\footnotemark\relax\let\c@footnote\@tmpcnta}}

\makeatother

\usepackage[inline]{enumitem}
\newlist{inlinelist}{enumerate*}{1}
\setlist[inlinelist]{label=(\roman*)}

\copyrightyear{2024}
\acmYear{2024}
\setcopyright{acmlicensed}\acmConference[SIGIR-AP '24]{Proceedings of the 2024 Annual International ACM SIGIR Conference on Research and Development in Information Retrieval in the Asia Pacific Region}{December 9--12, 2024}{Tokyo, Japan}
\acmBooktitle{Proceedings of the 2024 Annual International ACM SIGIR Conference on Research and Development in Information Retrieval in the Asia Pacific Region (SIGIR-AP '24), December 9--12, 2024, Tokyo, Japan}
\acmDOI{10.1145/3673791.3698424}
\acmISBN{979-8-4007-0724-7/24/12}
\begin{document}

\DeclareGraphicsExtensions{.png,.pdf}
\title{Offline Evaluation of Set-Based Text-to-Image Generation}
\author{Negar Arabzadeh}
\authornote{work done while at Google}
\orcid{0000-0002-4411-7089}
\affiliation{%
  \institution{University of Waterloo }
  \city{Waterloo}
  \country{Canada}
}
\email{narabzad@uwaterloo.ca}

\author{Fernando Diaz}
\orcid{0000-0003-2345-1288}
\affiliation{%
  \institution{Google}
  \city{Pittsburgh}
  \country{USA}
}
\email{diazf@acm.org}

\author{Junfeng He}
\orcid{0009-0004-5465-5659}
\affiliation{%
  \institution{Google}
  \city{Mountain View}
  \country{USA}
}
\email{junfenghe@google.com}

\settopmatter{printacmref=true}

\begin{abstract}
Text-to-Image (TTI) systems often support people during ideation, the early stages of a creative process when exposure to a broad set of relevant 
images can help explore the design space. 
Since ideation is an important subclass of TTI tasks, understanding how to quantitatively evaluate TTI systems according to how well they support ideation is crucial to promoting research and development for these users.  
However, existing evaluation metrics for TTI remain focused on distributional similarity metrics like Fréchet Inception Distance (FID).  We take an alternative approach and, based on established methods from ranking evaluation, develop TTI evaluation metrics with explicit models of how users browse and interact with sets of spatially arranged generated images.  Our proposed offline evaluation metrics for TTI not only capture how relevant  generated images are with respect to the user's ideation need but also take into consideration the diversity and arrangement of the set of generated images. We analyze our proposed family of TTI metrics using human studies on image grids generated by three different TTI systems based on subsets of the widely used benchmarks such as MS-COCO captions and Localized Narratives as well as  prompts used in naturalistic settings.  Our results demonstrate that grounding metrics in how people use systems is an important and understudied area of benchmark design.
\end{abstract}



\maketitle

\section{Introduction}
\label{sec:intro}

Generative models and specifically text-to-image (TTI) models are being increasingly used in creative tasks, especially through ideation, the process of generating a list of design candidates in the beginning of a creative design task \cite{koch:ai-ideation,karimi:ideation,yang2020evaluation,dang2022prompt,jeon:fashion-ideation}. In this paper, we study TTI systems, a class of models particularly well-suited for ideation in the visual arts \cite{ko:tti-visual-artist-ideation}, architecture \cite{paananen:tti-architecture-ideation}, and crafting \cite{vartiainen:tti-craft-ideation}.  
In order to robustly benchmark TTI systems, researchers need to develop both  datasets  (i.e., prompts and relevance information such as a target image) \cite{https://doi.org/10.48550/arxiv.2211.12112,yu2022scaling,saharia2022photorealistic} as well as \textit{evaluation metrics}\ \cite{frolov:tti-metrics,otani:tti-evaluation-status}. These metrics can either be based on human rater preferences between a pair of system outputs (e.g., through a side-by-side comparison for a fixed prompt) or  on offline metrics based on the automatic comparison of model output to an example system target.  

While there have been advances in the critique and mitigation of datasets involved in benchmarking TTI systems \cite{qadri:tti-representation,smith2023balancing},  evaluation metrics remain relatively immature and deserving of more attention \cite{otani:tti-evaluation-status}.  Evaluation methods based on manual side-by-side comparisons of system output \cite{arabzadeh2024comparison,arabzadeh2024fr,alaofi2024generative,arabzadeh2022shallow}, although effective at detecting differences between systems, suffer from multiple inefficiencies. On the one hand, this method incurs time costs arising from  recruiting and training participants as well as conducting the assessment alone; when operating in rapidly advancing environments, these time costs can limit the space of models evaluated (and developed). On the other hand, because   each pair of systems needs to be manually compared, the evaluation scales quadratically in the number of systems \cite{deckers2022infinite}; as a result, the financial cost of model evaluation may be prohibitive. Offline or automated evaluation metrics address these scalability issues by using information such as target image data to algorithmically score and compare systems \cite{heusel2017gans,salimans2016improved}.  The most popular approach is to use the Fréchet Inception Distance (FID) \cite{heusel2017gans}, a method that compares the distribution of generated images over a workload of prompts to the distribution of example or target images over the same workload of prompts.  FID, while mathematically convenient, is deficient in several ways \cite{https://doi.org/10.48550/arxiv.1911.07023,https://doi.org/10.48550/arxiv.2103.09396}.  First, FID compares population-level statistics and, therefore,  a system can generate a set  of images comparable to the target set while individually being poor responses to the prompt.  Second, because FID is a population-level metric, the benefits of per-prompt error analysis disappear.  This includes evaluating for sub-population metrics such as fairness.  Moreover, when comparing a pair of systems, because FID measures population level differences, evaluation cannot leverage matched statistical tests, known to improve sensitivity.  Finally, FID may inaccurately measure TTI effectiveness because it is computed using a collection of generated images within and across prompts without paying attention to how they may be used or interacted with by users. This risks metric divergence \cite{morgan:hill-climbing}, when there is a disconnection between an evaluation metric and the construct we are interested in measuring, a more general version of the value alignment problem \cite{wiener:alginment}. FID does not explicitly model user behavior or task, even though we know users engage with TTI systems in the context of ideation through a grid interface.  As a result, evaluation may be brittle, especially as systems become effective enough to be operating in a region of the performance landscape where FID cannot effectively detect differences (cf. similar phenomenon in search evaluation \cite{voorhees:too-many-relevants}).

%
Seeing the need for reusable offline evaluation as a necessary benchmarking tool, we develop novel TTI evaluation metrics based on explicit models of how users engage with system decisions.  This responds to a recent call to abandon metrics like FID that do not model human behavior and perception \cite{otani:tti-evaluation-status}.  We leverage theories from the ideation literature to understand what people need during ideation \cite{guilford:creativity,shah:ideation-metrics,kerne:curation-metrics,oppenlaender2022creativity} and techniques from Information Retrieval (IR) to formally model these concepts \cite{diaz:neurips-2020-tutorial}.  In particular, we adapt two well-studied IR metrics, expected reciprocal rank \cite{chapelle2009expected} and  rank-biased precision \cite{moffat:rbp}, to account for grid layouts,  generated images, and ideation intent.  \textcolor{black}{ We note that to leverage the soft relevance labels based on similarities to example images, in this study, we assume that  there is always one or more ground truth example images known to be relevant to the given prompt.}

We conduct experiments on comparing our proposed set of metrics with traditional ones, demonstrating that
\begin{inlinelist}
    \item modeling sequential user browsing increases consistency with human preferences,
    \item modeling novelty and variety increases consistency with human preferences, and
    \item modeling image visual saliency increases consistency with human preferences.
\end{inlinelist}
We observe this behavior on three different datasets including MS-COCO captions \cite{chen2015microsoft}, the localized narratives \cite{pont2020connecting} and a set of aggregated prompts generated by humans. We release our code at \url{https://github.com/Narabzad/Set-Based-Text-to-ImageGeneration}.

\section{Related work}
\textbf{Text to Image Generation Evaluation} Despite the recent advances in text guided image generation models \cite{ramesh2022hierarchical,radford2021learning,rombach2022high,yu2022scaling,saharia2022photorealistic,ramesh2021zero,ding2021cogview,gafni2022make,koh2021text,liao2022text,li2023gligen,wang2023imagen}, evaluating such systems have not been studied extensively. The lack of evaluation tools for such systems has been identified as a serious issue within the community \cite{pavlichenko2022best,otani:tti-evaluation-status,cho2023dall,luzi2023evaluating,otani2023toward}. 
The Inception Score (IS) measures properties of \textit{unconditional} generated images for
\begin{inlinelist}
    \item low-entropy across image classes, and
    \item high entropy across image classes over a population of generated images \cite{salimans2016improved}.  
\end{inlinelist}
However, the Inception Score (IS) is susceptible to gaming a perfect score by creating one image for each of the base embedding classes. Therefore, while better performance is possible, the metric does not fully capture the potential for improvement. 
Additionally, IS is appropriate for assessing only certain properties of a general, unconditional image generation model and is not suitable for TTI evaluation. As a mathematical refinement of IS, the Fréchet Inception Distance (FID) compares the distribution of Inception embeddings of real and generated examples \cite{heusel2017gans,chong2020effectively}. A lower FID indicates a higher similarity between the real and generated distributions, and thus more realistic images. However, like IS, FID was designed for unconditional image generation, making its use for TTI evaluation somewhat contrived \cite{arabzadeh2024fr}. Although other methods, like Kernel Inception Distance \cite{bikowski2021demystifying}, have proposed further mathematical improvements,  FID remains the dominant evaluation metric \cite{otani:tti-evaluation-status}.
Multiple studies have demonstrated that population-level metrics like FID are not correlated with human preferences \cite{liu2018improved,parmar:fid-brittle,otani:tti-evaluation-status}.  In particular, \citet{otani:tti-evaluation-status} rigorously   analyzed  TTI evaluation, finding weak alignment with human behavior and perception as a primary cause of this inconsistency.

A second group of metrics evaluates the alignment between generated images and the user prompt. Some metrics achieve this by algorithmically captioning the generated image and then computing text similarity to the user prompt 
\cite{wang2021simvlm,DBLP:journals/corr/abs-2202-04053}. Other metrics measure semantic object accuracy, which assesses how well objects in the generated image match terms in the caption \cite{hinz2020semantic,yarom2024you,hu2023tifa}. This group of evaluation metrics is limited by the performance of a second model (e.g., caption generation, object detection model, or visual question answering models) and usually does not show a high correlation with human-generated prompts, as reported in previous works \cite{saharia2022photorealistic,yu2022scaling,DBLP:journals/corr/abs-2202-04053,park2021benchmark}.
In addition, they lack the flexibility to be applied to \textit{set-based} evaluation of TTI systems, which is the focus of our work. Even state-of-the-art caption generation models struggle to produce distinct captions for images generated from the same TTI system for a given prompt \cite{sumbul2020sd,kanani2020improving,gao2022caponimage}. Our experiments demonstrate that even with a SOTA  pretrained image captioner model such as VL-T5\cite{cho2021unifying}, there are significant overlaps between the generated captions for images generated from an individual TTI system for a single prompts. Consequently, the evaluation metric would yield the same relevance score for all generated images within the set. As a result, these metrics are not effective in evaluating TTI systems on set-based instances, which is the focus of our work. In a recent study \cite{xu2024imagereward}, the authors introduced a method to score and evaluate human preferences. Similarly, in \cite{kirstain2023pick} the authors collected a preference-based dataset to train a CLIP-based scoring function so called as PickScore to predict human preferences on individual pair of generated images for a given prompt.
However, their focus was on individual generated images rather than considering set-level evaluation, which crucially incorporates diversity alongside relevance. In our work, we only focus on TTI evaluation metrics that are applicable to set-based TTI generation evaluation.

There have been also attempts to develop evaluation pipelines for TTI models \cite{bakr2023hrs,huang2023t2i,kirstain2023pick,hu2023tifa}. For instance, \citet{https://doi.org/10.48550/arxiv.2211.12112} created a set of fifty challenging tasks and applications for state-of-the-art TTI models. 
 As a result, several curated benchmarks for TTI evaluation have been developed. Moreover \citet{yu2022scaling} introduced PartiPrompts, a comprehensive benchmark of 1,600 English prompts covering 12 different categories and 11 challenge aspects. In PartiPrompts, similar to \citet{xu2024imagereward} human annotators were asked to select the better generated image from two models (one image per model).
Although the majority of preference-based TTI evaluation focuses on pairs of single images, \citet{saharia2022photorealistic} conducted preference-based evaluations between sets of  images generated by two models. To the best of our knowledge, their collected dataset is the only work that conducted evaluation on set of images. In our work, we follow the same concept and aim to conduct experiments comparing sets of generated images, albeit on a larger scale.


\textbf{Diversity}
Measuring the diversity of image sets has a long history in multimedia retrieval \cite{van2008diversifying,maisonnasse2009lig,ah2008xrce}. For example, the MediaEval 2017 benchmark introduced image search result diversification in the context of photo retrieval \cite{zaharieva2017retrieving}, arguing that diversity within image search results is crucial because it
\begin{inlinelist}
    \item addresses the needs of different users,
    \item tackles requests with unclear information needs, 
    \item widens the pool of possible results for increasing the performance, and
    \item reduces redundancy in the results.
\end{inlinelist}
We adopt similar desiderata motivated from the the precise ideation context in which TTI is often used.
While MediaEval could use relevance and diversity unmodified IR metrics because images were explicitly rated, TTI needs to impute these ratings because images are generated.


\section{Method}
\subsection{Problem Definition}
\label{problemdef}

\begin{figure*}[t!]
\centering
\vspace{-1em}
    \includegraphics[clip, trim=0cm 3cm 5cm 3.8cm, scale=0.8]{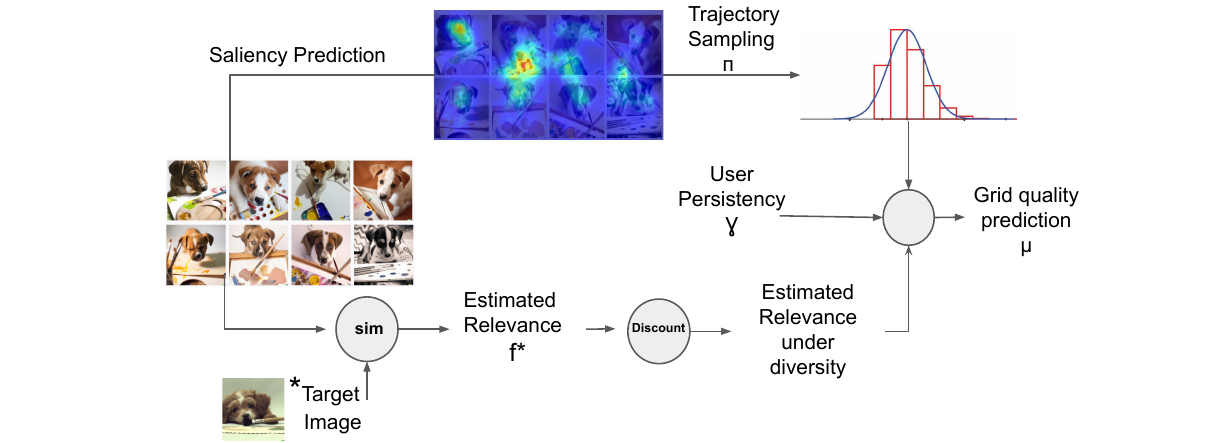}
\vspace{-1em}
\caption{\textcolor{black}{Overview of our proposed approach }}
\vspace{-1em}
\label{fig:overview}
\end{figure*}


We define the \textit{TTI Task} as: Given a prompt or query $\prompt$, a TTI system $\system$, generates an $\gridWidth\times\gridHeight$ matrix $\images$ where each element is an image.  
If the TTI interface returns a ranked list of length $k$, we generate a $1\times k$ matrix and similarly when a grid is required, the TTI system presents the generated images in a grid view of $m \times n$ images.  
Further, we define the \textit{TTI Evaluation Task} as: An evaluation metric $\evalmetric$ is a function that, given an arrangement of generated images $\images$, a prompt $\prompt$, and side information $\promptutility$ about the image utility (e.g. an example target image), computes a scalar value where a higher value indicates better performance of system.
We summarize the performance of system $\system$ over a space of prompts $Q$ with the  expected value of $\evalmetric$ over the distribution of queries in $Q$, $\mathbb{E}_{ \prompt \sim Q}[ \evalq ]$.
 



\subsection{Designing a Metric for TTI}
\label{proposedmetric}
As mentioned in Section \ref{sec:intro}, in the context of creative design, users leverage TTI systems to generate design candidates to support ideation. Dominant theories for measuring the quality of an ideation process focus on four criteria \cite{guilford:creativity,shah:ideation-metrics,kerne:curation-metrics}. \textit{Fluency} refers to the total number of relevant items generated. \textit{Variety} refers to the number of unique types of relevant items generated. \textit{Novelty} refers to how different relevant items are from all previously generated items. Finally, \textit{quality} refers to the degree of relevance of generated items. We are interested in designing metrics that capture these different dimensions of ideation effectiveness.Here,  we will review relevant concepts from information retrieval evaluation as they pertain to TTI evaluation.

\textbf{A General Model of User Behavior:}
Let $\images$ be the matrix of $\numgen=\gridWidth\gridHeight$ images generated for a specific prompt, where we index each generated image from 1 to $\numgen$.  When inspecting a grid, we define the trajectory $\trajectory$ as be the specific sequence of inspected images represented as a permutation of $[1,\numgen]$, where $\trajectory_i$  refers to the $i$th examined image index.  We  begin our metric development in the context of classic IR evaluation where users inspect a one-dimensional ranked list in a serial, deterministic order.  We will return to grid-based interfaces at the end of this section.

We also have available information about the \textit{relevance} of an image to the user for a prompt $\prompt$.  Let $\relevance(\image)\in[0,1]$ represent the relevance of an image $\image$ to the user.  

Given a trajectory $\trajectory$, we define a family of metrics based on how the user might engage with the images in linear order.  The \textit{position-based model} models the probability of a user inspecting the image at rank position $i$ in the trajectory as $\patience^{i-1}$, where $\patience\in[0,1]$ is a free parameter controlling the depth the user is likely to reach \cite{craswell2008experimental,richardson2007predicting}. 
The \textit{cascade model} models a user who, after inspecting a relevant image, might be satisfied and terminate their scan \cite{chapelle2009dynamic,craswell2008experimental,chapelle2009expected}.  Let the probability that a user is satisfied by an image be a monotonically increasing function of its relevance, $\satiation(\relevance(\image))$, which we represent as shorthand using $\satiation(\image)$ for clarity.  In this case, the probability that they reach rank position $i$ is $\patience^{j-1}\prod_{j=1}^{i-1}(1-\satiation(\trajectory_j))$.
In the remainder of this section, we will demonstrate how to use these models to design metrics capturing fluency, quality, novelty, and variety.

\begin{table*}[]
\centering
\caption{Examples of prompts from $\MSCOCO$ ,$\LNCOCO$ and $\prompts$ dataset as well as average and standard deviation $\sigma$ of length of prompts in each dataset.   }
\label{fig:example}
\scalebox{1}{
\begin{tabular}{p{1.1cm}cp{0.95cm}cp{9cm}}
\textbf{Dataset} & \textbf{\#} & \textbf{\#Words} & \textbf{$\sigma$} &  \textbf{Example Prompt} \\ \hline
\MSCOCO & 500 & 10.3  & 2.2 &
 A herd of cows standing on a grass covered hillside. 
 \\ \hline
\LNCOCO & 500 & 42.8  & 19.2 &  In this picture we can see three cows standing on the grass. There is a tree and few mountains are visible in  the background. 
    \\ \hline
\prompts & 500 & 11.6  & 9.4 & Origami cow flying over the moon.  \\ 
\end{tabular}}
\vspace{-1em}

\end{table*}

\textbf{Fluency:}
To measure fluency, we  define a metric based on the expected number of relevant images the user will see.  For the position-based model, rank-biased precision or RBP \cite{moffat:rbp}, is defined as,
\begin{align}
\label{rbp}
    \rbp(\trajectory)&=\sum_{i=1}^\numgen \relevance(\trajectory_i)\patience^{i-1}
\end{align}
Under the cascade model, extended expected reciprocal rank (ERR) metric \cite{chapelle2009expected}, is defined as,
\begin{align}
    \label{err}
    \err(\trajectory)&=\sum_{i=1}^\numgen \relevance(\trajectory_i)\patience^{i-1}\prod_{j=1}^{i-1}(1-\satiation(\trajectory_j))
\end{align}

\textbf{Quality:}
\label{prior}
Unlike classic IR evaluation, we cannot  \textit{a priori} judge the relevance of all possible images, we leverage a wide variety of side information  to estimate the relevance of generated images.  
In general, this could include information about user preferences, image attributes, or anything else helpful to estimate relevance.  In our study, we assume that we have access to one or more example images known to be relevant to the prompt.  
Given a generated image $\image$ and a relevant image $\targetimage$, we estimate the relevance of $\image$ as the following where $\imageembedding$ is an image embedding function, for example based on the activation of an interior layer of a neural network:
\begin{align}
\relevance(\image) &= \similarity{\imageembedding(\targetimage)}{\imageembedding(\generatedimage)}
\end{align}
\citet{carterette:similarity-evaluation} uses a similar approach to label unlabeled documents in classic IR evaluation. 

\textbf{Variety and Novelty:}
 In TTI evaluation, for a given prompt, we only have a few example images and therefore do not have information to measure novelty of a generated image to the user or in general.  However, we can measure the novelty of an image to the user \textit{while they scan the list}.  Given the similarity of novelty within a list to variety within a list, we consider both of these criteria together.  In situations where we have additional information about the novelty of an image to a user or in general, we suggest these criteria be decoupled. 
 In order to measure novelty within a list, we leverage existing methods designed to measure diversity.  Similar to Maximum Marginal Relevance \cite{guo2010probabilistic}, we model novelty by discounting the relevance of an image based on previously seen images.  The discount factor for the $i$th image in $\trajectory$ is defined as the following where $\novelty(i,\trajectory)$  can be used to discount $\relevance(\trajectory_i)$ which we can then use in either $\rbp$ or $\err$.  
\begin{align}
\label{discount}
    \novelty(i,\trajectory)&=1-\max_{j\in[1,i-1]}\similarity{\imageembedding(\trajectory_i)}{\imageembedding(\trajectory_j)}
\end{align}

\textbf{Expected Metrics over Trajectories:}
\label{trajectory}
In more general grid-based interfaces, we cannot assume that users will all follow a deterministic trajectory when presented with an arrangement of images  \cite{fdiaz:robust-mouse-tracking-models}.  Users may inspect images in arbitrary trajectories based on their position and attractiveness.  As such, we can compute 
the expected metric value over all possible trajectories. Let $\prob(\trajectory)$ be the probability that the user scans the images in the order represented by $\trajectory$; in other words, $\prob(\trajectory)$ is a multinomial distribution over all permutations in $S_\numgen$.
Although a na\"{i}ve model might consider a uniform distribution over trajectories, we know that users tend to be more attracted to certain images based on their position and visual features \cite{fdiaz:robust-mouse-tracking-models}. Inspired by the observation that users tend to look at the most salient images first \cite{navalpakkam:choice-salience,silvennoinen2016appraisals}, we model $\prob(\trajectory)$ using a Plackett-Luce model based on predicted image salience \cite{plackett1975analysis}. Let $\salience(\image)$ be the visual salience of $\image$ represented as a positive scalar value. After normalizing this value over all images to sum to one, we iteratively sample an image from this multinomial \textit{without replacement}, generating a sequence of images (i.e. $\trajectory$). This process can be executed efficiently by using the Gumbel-softmax method \cite{bruch:stochastic-ltr}.Figure \ref{fig:overview} displays a general overview of our proposed approach for evalauting TTI systems.

\begin{figure*}
    \includegraphics[clip, trim=0cm 1cm 1cm 0cm, scale=0.6]{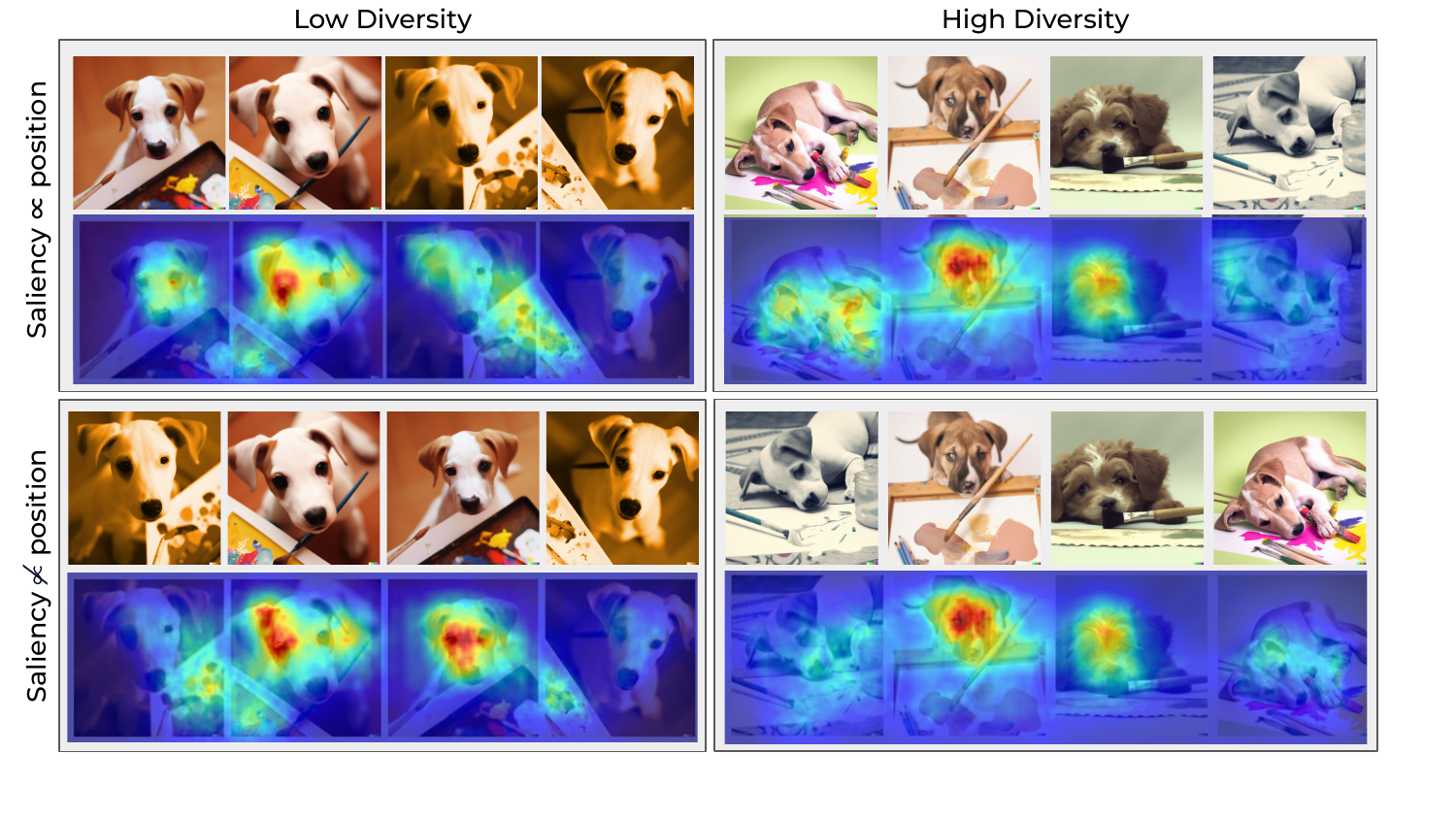}
    \vspace{-1em}
    \caption{
    Four image grids and associated salience maps for the prompt `a cute puppy is painting'.   Images grids in the left column have lower diversity than those in the right column as reflected by the different positions and breeds on puppies.  In image grids in the top row, images tend decrease in salience from left to right; in the bottom row, there is less of a relationship between position and salience.
}  
\label{fig:main}
\end{figure*}
\section{Experimental Setup}
\subsection{TTI Systems}
\textcolor{black}{To run experiments and validate how our proposed evaluation metrics can distinguish between the two sets of generated images, we took three  state-of-the-art TTI systems \cite{saharia2022photorealistic,yu2022scaling}}.  
In order to preserve the anonymity of the systems and emphasize evaluation, in this paper, we refer to these as systems \texttt{A}, \texttt{B} and \texttt{B'}. \texttt{B'} is the smaller version of system \texttt{B}, with fewer number of parameters. In prior work, systems \texttt{A} and \texttt{B} have shown competitive performance and consequently, the smaller version of system \texttt{B}, system \texttt{B'}, has shown relatively worse performance compared to system \texttt{A}. The assumption that a model with fewer parameters has worse performance compared to a larger model of the same class has  previously been used for evaluation purposes \cite{pavlichenko2022best}. 

\subsection{Datasets}
To conduct our analysis, we require sets of prompts with associated prior information. As discussed in Section \ref{prior}, each prompt is linked to at least one relevant target image. 
We utilized the COCO dataset \cite{lin2014microsoft}, which provides two distinct representations of information needs: 1) COCO captions ($\MSCOCO$) \cite{chen2015microsoft} and 2) Localized Narratives ($\LNCOCO$) \cite{pont2020connecting}. Additionally, we collected a third dataset, $\prompts$, comprising of prompts entered by users of a TTI demo.

The examples in Table \ref{fig:example} highlight the difference between prompts used for these experiments for the same image. As shown in the first two rows of this Table, the same image was described with 10 words when using $\MSCOCO$ captions and 24 words when using $\LNCOCO$ as prompts. We demonstrate how a same image was described with 10 words when using $\MSCOCO$ captions and 24 words when using $\LNCOCO$ as prompts. 
We are interested in investigating whether our proposed metrics and baselines are robust with respect to the lengths of the prompts for the same prior information. While the average number of words in $\prompts$ and $\MSCOCO$ seem close (11.6 vs. 10.3), the standard deviation of the number of words in the prompts of these two datasets differs significantly. The standard deviation of the number of terms in $\prompts$ is 9.41, while this number for $\MSCOCO$ is only 2.21. From Table \ref{fig:example}, we can see that prompts in the $\MSCOCO$ dataset are consistently short, while in the $\prompts$ dataset, the length of the prompts varies more.
We conclude that prompts in the $\MSCOCO$ dataset are consistently short, while in the $\prompts$ dataset, the length of the prompts varies more.

\textbf{COCO Captions}
\label{MSCOCO}
The COCO dataset is widely used to evaluate various deep learning and computer vision tasks, including TTI systems \cite{yu2022scaling,saharia2022photorealistic,ding2021cogview,sharma2018chatpainter,chen2022re,DBLP:journals/corr/abs-2111-13792,lin2014microsoft}.
The MS-COCO captions dataset contains over 200K images, each annotated with five captions. For this paper, we randomly sampled 500 images from the MS COCO 2017 validation set. For each image, we randomly selected one of the captions as the input prompt for experiments and used the associated image as the target image.

\textbf{Localized Narratives}
Recent work has raised concerns that COCO captions might not be realistic surrogates for human prompts
\cite{yu2022scaling}. Therefore, 
we use
Localized Narratives dataset which are a subset of the MS-COCO dataset, and their associated text is four times longer than COCO captions on average \cite{pont2020connecting}. Inspired by previous work \cite{koh2021text,DBLP:journals/corr/abs-2111-13792,zhang2021cross}, we also conducted experiments with generated images from more detailed descriptions of images (localized narrations) and examined how the longer version of prompts affects the performance of TTI systems.


\textbf{Prompts}
Describing an image through captions or detailed narrations is different from issuing a prompt to address an information need in various contexts. To ensure that our evaluation metric is representative of prompts that appear in practice, we also analyzed performance on human-generated prompts.
We collected a third dataset including 1,500 prompts from real users, referred to as $\prompts$, consisting of prompts entered by users of a TTI demo. This dataset is more representative of the types of prompts that TTI systems will encounter in practice, and they cover a wide range of topics and styles. Some examples of prompts in $\prompts$ are \textit{"A futuristic city with lots of greenery", "A small cabin in the woods"}, and \textit{"A cozy reading nook in a library".} 
The TTI demo was used for generating images for various purposes, including research, design, and presentations. We also recorded the generated grids from two TTI systems, \texttt{A} and \texttt{B}, and collected feedback on individual images, including positive and negative signals, such as thumbs up and down, as well as implicit signals, such as downloads, copies, or shares.
We used both implicit and explicit positive feedback to create a set of target images for each prompt, selecting only those prompts that were tried on both TTI systems and received positive feedback on both. This resulted in a set of target images, consisting of at least one image from each of the two TTI systems, for each prompt. We randomly sampled 500 such prompts that satisfied these requirements.

\subsection{Metric Variations}
We present our six different  metric variations  in Table \ref{tab:metrics}.  $\rbp$ and $\err$ metrics are defined in Equation \ref{rbp} and \ref{err} respectively. The subscripts $\novelty$ discounts the relevance function as  explained in Equation \ref{discount}. When $\trajectory$ is sampled based on uniform distribution the metric has a superscript $u$ otherwise the $\trajectory$ is sampled using the normalized image saliency and unless specified, we adopt a value of $\gamma$ equal to 0.9, as suggested in the literature \cite{carterette2011system}. We compute the average of the metric over 100 sample trajectories for each prompt. The similarity between generated images and target images is calculated using the cosine similarity of the embedded vectors of images obtained from Inception V3 which has been widely used in downstream vision tasks and evaluation methodologies \cite{szegedy2016rethinking,ramesh2022hierarchical,xia2017inception,wang2019pulmonary,lin2019transfer,barratt2018note,obukhov2020quality}.

\begin{table*}[]
\centering
\caption{Different variations of our proposed metrics}
\scalebox{1}{
\begin{tabular}{ccccccccc}
Metric & User Model & Relevance & $\prob(\trajectory)$ & & Metric & User Model & Relevance & $\prob(\trajectory)$\\ \cline{1-4} \cline{6-9}
$\rbp$ & Position-based & $\relevance(\image)$ & Saliency & & $\err$ & Cascade-based & $\relevance(\image)$ & Saliency  \\ 
$\urbp$ & Position-based & $\relevance(\image) \times \novelty(i,\trajectory)$ & Uniform & & $\uerr$ & Cascade-based & $\relevance(\image) \times \novelty(i,\trajectory)$ & Uniform \\
$\novrbp$ & Position-based & $\relevance(\image) \times \novelty(i,\trajectory)$ & Saliency & & $\noverr$ & Cascade-based & $\relevance(\image) \times \novelty(i,\trajectory)$ & Saliency \\
\end{tabular}}
\label{tab:metrics}
\end{table*}

\subsection{Baseline TTI Evaluation Metrics}
In addition to FID, we develop a simple diversity-based metric, $Diversity$, to assess the set-based diversity of TTI output. $Diversity$ measures the average pairwise similarity between all generated images. A higher pairwise similarity rate within a grid indicates \textit{lower} diversity. We compute pairwise similarity using cosine similarity of the embedded vectors of images obtained from Inception V3.

\subsection{Modeling Salience}
Although there are no existing saliency models or gaze data available for a grid of images, we can use models trained on webpages, which usually contain multiple images (or even a grid of images sometimes). There are several existing free viewing gaze data on webpages \cite{stonybrook, fiwi}, and among them, \cite{stonybrook} is the latest and largest, with gaze data on 450 webpage screenshots, collected from 41 people with eye trackers. 
450 webpages are a relatively small data set, and insufficient to train the webpage saliency model. We follow the typical training paradigm in saliency modeling area: first train the model with Salicon data \cite{jiang2015salicon}, a large scale saliency data set on 10K natural images, then fine-tune it with webpage gaze data. The loss functions we use is a combination of KLD and NSS \cite{bylinskii2018different}, both of which are popular loss function for saliency models.  Our webpage saliency architecture essentially follows the SimpleNet idea in \cite{simplenet} with some modification and simplification. More specifically it takes a MobileNet V3 (pretrained on ImageNet classification data) as backbone, and extracts embeddings from 4 layers of conv 2, 4, 6, 8. On each embedding, we apply two conv layers, where the first layer has a kernel $3\times3$, number of channels matching the input embedding, max pooling 3 and relu operator and the second layer has a kernel $1\times1$, 1 channel and relu operator. Then the output of each branch is converted back to the resolution of the input with a bilinear resize operator. The outputs of the 4 branches are then summed to 1 channel, followed up by a sigmoid function to generate the output saliency map. 
Finally, we average the predicted saliency of individual images within the grid and use it as initial distribution to sample $\trajectory$ \footnote{We release our code at \url{https://github.com/Narabzad/Set-Based-Text-to-ImageGeneration}}.

\subsection{Data Annotation}
\subsubsection{Annotation Process}
We curated a dataset of human preferences over sets of generated images for prompts from the three above-mentioned datasets.   An annotation study was conducted using an internal platform, where annotators selected their preferred set of images for each prompt.
 Based on prior work in task-based evaluation of information retrieval systems \cite{Borlund:1997ul}, we grounded the annotation in  a simulated work task  provided to annotators as,

\textit{``
You and your coworkers are trying to come up with an image for a project presentation. Together, you all have come up with a description of the image, which we will refer to as a `prompt' (shown in blue below). Two designers, X and Y, have sketched possible images for the group to decide on the images to use. Which designer’s sketches would you prefer to present to the group to decide on the presentation image?''}

The instructions, along with a prompt, were displayed at the top of the assessment interface. Two sets of eight images were presented in a $2 \times 4$ grid view, one from TTI system X on the left and one from another TTI system Y on the right, immediately below the guidelines and the prompt. Annotators rated which of the grids they preferred on a five-point Likert scale: 1) X is much better than Y, 2) X is somewhat better than Y, 3) X and Y are the same, 4) Y is somewhat better than X, and 5) Y is much better than X.

To ensure the quality of the collected annotations, each pair of sets of generated images was annotated three times with different annotators. We used the consensus agreement among all three judgments to generate the final annotation results. We ran preference-based annotation between each pair of systems, i.e., \texttt{A} vs. \texttt{B}, \texttt{A} vs. \texttt{B'}, and \texttt{B} vs. \texttt{B'}, on 500 prompts from each of the $\MSCOCO$, $\LNCOCO$, and $\prompts$ datasets.

In total, we collected 13,500 annotations for our dataset. Additionally, we conducted 100 quality check tests in which we presented a generated set of images for a given prompt and a random set of generated images from a different prompt. We assumed that the randomly generated set of images would always be less relevant. We are pleased to report that all of the quality tests were passed with the consensus results. Furthermore, the crowd sourcing platform we utilized had their own internal quality checks and qualification tests which were passed at the most satisfactory level. On average, each annotation took 106 seconds to complete and although we paid the raters hourly, on average we paid 0.50 USD per annotation.

\begin{figure*}[t!]
\centering

\includegraphics[clip, trim=5.15cm 4.1cm 5.9cm 2.6cm,scale=0.8]{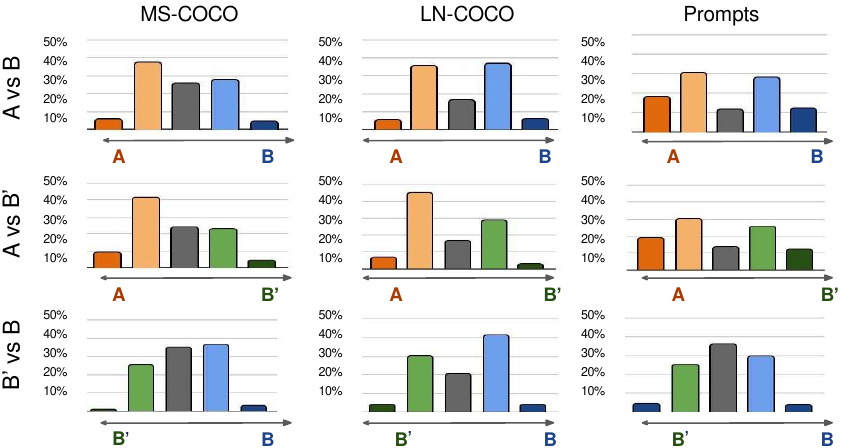}
\caption{{
Annotation results of  TTI
systems \texttt{A} (orange), \texttt{B} (blue)  and \texttt{B'} (green)
on the three datasets. Given the TTI systems X/Y are on the left/right side of the arrows beneath a  sub-figure, the bars in the sub-figure  present if "X is much better than Y", "X is somewhat better than Y", "X and Y are the same", "Y is better than X" and "Y is much better than X" consecutively. }}
\label{fig:annotation}
\end{figure*}
\subsubsection{Agreement Rate}
Three different annotators judged each pair of sets of generated images.  We computed Fleiss' $\kappa$ independently for each dataset. 
In Table \ref{agreement}, we provide the agreement rate between the annotators when annotating different pairs of systems on the three datasets separately, as well as the total agreement on all the annotated pairs of grids. 
We also report the agreement rate on a three-point scale where we collapse ratings in the same \textit{direction} (e.g., considering `A is much better than B' and `A is somewhat better than B' as one option `A is better than B'). As shown in this Table, the agreement rate between the three annotators was quite promising, and the majority of the prompts received at least two out of the three annotator's agreement. Even on a 5-scale consensus result, which is harder to achieve higher agreement on, only 13.6\% of the prompts from $\MSCOCO$, 12.3\% of prompts from $\LNCOCO$, and 12.6\% of prompts from $\prompts$ dataset did not have any consensus agreement between the three annotators. We qualitatively observed some of the prompts that received 0 agreement rate between the three annotators and noted that in these special cases, both of the models generated a relevant set of images, and thus the annotator's selection became less robust.

When looking at agreement across annotators for grades on our Likert scale, we observe moderate agreement $\MSCOCO$ ($\kappa=0.57$) and $\LNCOCO$ ($\kappa=0.58$) and fair agreement for $\prompts$ ($\kappa=0.33$).  This suggests that systems are easier to precisely distinguish in $\MSCOCO$ and $\LNCOCO$ while the $\prompts$ dataset is more challenging for annotators.  We also computed agreement in \textit{direction}, collapsing the bottom two and top two grades.  When computing the directional agreement, we observe substantial agreement for $\LNCOCO$ ($\kappa=0.62$) and moderate agreement for $\MSCOCO$ ($\kappa=0.56$)  and  $\prompts$ ($\kappa=0.41$).  We also observed that, despite perhaps being more challenging when evaluating these specific systems, the $\prompts$ dataset has equal or better majority agreement compared to $\MSCOCO$ and $\LNCOCO$.  From an evaluation perspective, the difficulty in distinguishing systems could arise from, for example, harder to distinguish pairs of image sets or more ambiguous prompts.  If we had observed substantially lower  agreement, we would question whether these systems were in fact different in terms of the prompts in the $\prompts$ dataset.  

Additionally, by comparing the agreement rate on the three datasets, we observe the agreement rate on the $\prompts$ dataset is significantly lower than the other two datasets. For instance, while 57.8\% of the prompts in the $\MSCOCO$ dataset and 61.8\% of prompts in the $\LNCOCO$ dataset had 3 out of 3 consensus agreement on a 5-scale annotation result, only 28.5\% of the prompts in the $\prompts$ dataset had full agreement rate between the 3 annotators. Obviously, the agreement rate increases when we only consider the direction of the options and neglect the magnitude. For instance, the 3/3 agreement rate increases from 57.8\% to 63.5\% for $\MSCOCO$, 61.8\% to 68.6\% for $\LNCOCO$, and from 28.5\% to 45.8\% in the $\prompts$ dataset. However, we still see a notable difference between the three datasets in the agreement rate. We hypothesize that this difference could be because the prompts in the $\prompts$ dataset are more challenging and point to more specific concepts and design patterns, such as the vibes, the quality, style, etc. However, the prompts in the $\LNCOCO$ and $\MSCOCO$ datasets do not elaborate on such specifications, and thus they obtained a higher agreement rate between the three annotators.

\begin{table*}[!t]
\centering
\vspace{-1em}

\caption{Agreement rate between the three annotators on each of the datasets and between each pairs of TTI systems on a 5-level and 3-level Likert scale. Fleiss’ $\kappa$  is reported on both 3-levels and 5-levels consensus.}
\vspace{-1em}

\scalebox{1}{
\label{agreement}
\begin{tabular}{llcccccccc}
&  & \multicolumn{4}{c}{5-levels Consensus} & \multicolumn{4}{c}{3-levels Consensus}  \\
 & \multicolumn{1}{l}{Dataset} & \multicolumn{1}{l}{0/3} & 2/3 & 3/3 & \multicolumn{1}{c|}{ $\kappa$}& 0/3 & 2/3 & 3/3 & \multicolumn{1}{c}{ $\kappa$}  \\ \hline
\rowcolor[HTML]{DADADA} 
\multicolumn{1}{l}{\cellcolor[HTML]{DADADA}} & \multicolumn{1}{l|}{\cellcolor[HTML]{DADADA}$\MSCOCO$} & \multicolumn{1}{r}{\cellcolor[HTML]{DADADA}15.6\%} & \multicolumn{1}{r}{\cellcolor[HTML]{DADADA}24.5\%} & \multicolumn{1}{r}{\cellcolor[HTML]{DADADA}59.9\%} 
& \multicolumn{1}{c|}{-}&
\multicolumn{1}{r}{\cellcolor[HTML]{DADADA}13.7\%} & \multicolumn{1}{r}{\cellcolor[HTML]{DADADA}21.7\%} & \multicolumn{1}{r}{\cellcolor[HTML]{DADADA}64.6\%} 
& -
\\ \cline{2-10} 
\rowcolor[HTML]{DADADA} 
\multicolumn{1}{l}{\cellcolor[HTML]{DADADA}} & \multicolumn{1}{l|}{\cellcolor[HTML]{DADADA}$\LNCOCO$} & \multicolumn{1}{r}{\cellcolor[HTML]{DADADA}10.6\%} & \multicolumn{1}{r}{\cellcolor[HTML]{DADADA}26.9\%} & \multicolumn{1}{r}{\cellcolor[HTML]{DADADA}62.6\%} &
\multicolumn{1}{c|}{-} &
\multicolumn{1}{r}{\cellcolor[HTML]{DADADA}9.3\%} & \multicolumn{1}{r}{\cellcolor[HTML]{DADADA}18.9\%} & \multicolumn{1}{r}{\cellcolor[HTML]{DADADA}71.8\%} 
& -
\\ \cline{2-10} 
\rowcolor[HTML]{DADADA} 
\multicolumn{1}{l}{\multirow{-3}{*}{\cellcolor[HTML]{DADADA}\begin{tabular}[c]{@{}l@{}}A\\ vs\\ B\end{tabular}}} & \multicolumn{1}{l|}{\cellcolor[HTML]{DADADA}$\prompts$} & \multicolumn{1}{r}{\cellcolor[HTML]{DADADA}11.8\%} & \multicolumn{1}{r}{\cellcolor[HTML]{DADADA}60.7\%} & \multicolumn{1}{r}{\cellcolor[HTML]{DADADA}27.5\%}
&  \multicolumn{1}{c|}{-} &
\multicolumn{1}{r}{\cellcolor[HTML]{DADADA}4.0\%} & \multicolumn{1}{r}{\cellcolor[HTML]{DADADA}44.3\%} & \multicolumn{1}{r}{\cellcolor[HTML]{DADADA}51.7\%} 
& - \\ \hline
\multicolumn{1}{l}{} & \multicolumn{1}{l|}{$\MSCOCO$} & \multicolumn{1}{r}{12.6\%} & \multicolumn{1}{r}{25.7\%} & \multicolumn{1}{r}{61.7\%} 
& \multicolumn{1}{c|}{-} &
\multicolumn{1}{r}{10.8\%} & \multicolumn{1}{r}{23.8\%} & \multicolumn{1}{r}{65.4\%} 
& - \\  \cline{2-10} 
\multicolumn{1}{l}{} & \multicolumn{1}{l|}{$\LNCOCO$} & \multicolumn{1}{r}{12.7\%} & \multicolumn{1}{r}{25.4\%} & \multicolumn{1}{r}{61.8\%}
& \multicolumn{1}{c|}{-} & 
\multicolumn{1}{r}{10.1\%} & \multicolumn{1}{r}{21.5\%} & \multicolumn{1}{r}{68.4\%} 
& -
\\  \cline{2-10} 
\multicolumn{1}{l}{\multirow{-3}{*}{\begin{tabular}[c]{@{}l@{}}A\\ vs\\ B'\end{tabular}}} & \multicolumn{1}{l|}{$\prompts$} & \multicolumn{1}{r}{12.2\%} & \multicolumn{1}{r}{59.5\%} & \multicolumn{1}{r}{28.3\%} 
& \multicolumn{1}{c|}{-} &
\multicolumn{1}{r}{4.6\%} 
& \multicolumn{1}{r}{43.7\%} & \multicolumn{1}{r}{51.7\%} 
& - \\ \hline
\rowcolor[HTML]{DADADA} 
\multicolumn{1}{l}{\cellcolor[HTML]{DADADA}} & \multicolumn{1}{l|}{\cellcolor[HTML]{DADADA}$\MSCOCO$} & \multicolumn{1}{r}{\cellcolor[HTML]{DADADA}12.6\%} & \multicolumn{1}{r}{\cellcolor[HTML]{DADADA}25.7\%} & \multicolumn{1}{r}{\cellcolor[HTML]{DADADA}61.7\%} & 
\multicolumn{1}{c|}{-} &
\multicolumn{1}{r}{\cellcolor[HTML]{DADADA}18.2\%} & \multicolumn{1}{r}{\cellcolor[HTML]{DADADA}21.4\%} & \multicolumn{1}{r}{\cellcolor[HTML]{DADADA}60.4\%} 
& -
\\  \cline{2-10} 
\rowcolor[HTML]{DADADA} 
\multicolumn{1}{l}{\cellcolor[HTML]{DADADA}} & \multicolumn{1}{l|}{\cellcolor[HTML]{DADADA}$\LNCOCO$} & \multicolumn{1}{r}{\cellcolor[HTML]{DADADA}13.7\%} & \multicolumn{1}{r}{\cellcolor[HTML]{DADADA}25.2\%} & \multicolumn{1}{r}{\cellcolor[HTML]{DADADA}61.1\%} 
& \multicolumn{1}{c|}{-} &
\multicolumn{1}{r}{\cellcolor[HTML]{DADADA}12.4\%} &
\multicolumn{1}{r}{\cellcolor[HTML]{DADADA}22.1\%} & \multicolumn{1}{r}{\cellcolor[HTML]{DADADA}65.5\%} 
& -
\\  \cline{2-10} 
\rowcolor[HTML]{DADADA} 
\multicolumn{1}{l}{\multirow{-3}{*}{\cellcolor[HTML]{DADADA}\begin{tabular}[c]{@{}l@{}}B\\ vs\\ B'\end{tabular}}} & \multicolumn{1}{l|}{\cellcolor[HTML]{DADADA}$\prompts$} & \multicolumn{1}{r}{\cellcolor[HTML]{DADADA}13.6\%} & \multicolumn{1}{r}{\cellcolor[HTML]{DADADA}56.7\%} & \multicolumn{1}{r}{\cellcolor[HTML]{DADADA}29.7\%} 
& \multicolumn{1}{c|}{-} &
\multicolumn{1}{r}{\cellcolor[HTML]{DADADA}8.0\%} & \multicolumn{1}{r}{\cellcolor[HTML]{DADADA}58.1\%} & \multicolumn{1}{r}{\cellcolor[HTML]{DADADA}33.9\%} 
& - \\ \hline
\multicolumn{1}{l}{} & \multicolumn{1}{l|}{$\MSCOCO$} & \multicolumn{1}{r}{13.6\%} & \multicolumn{1}{r}{25.3\%} & \multicolumn{1}{r}{61.1\%} &
\multicolumn{1}{l|}{0.57}
& \multicolumn{1}{r}{14.2\%} & \multicolumn{1}{r}{22.3\%} & \multicolumn{1}{r}{63.5\%} 
& 0.56
\\  \cline{2-10} 
\multicolumn{1}{l}{} & \multicolumn{1}{l|}{$\LNCOCO$} & \multicolumn{1}{r}{12.3\%} & \multicolumn{1}{r}{25.8\%} & \multicolumn{1}{r}{61.8\%} & \multicolumn{1}{l|}{0.58} &
\multicolumn{1}{r}{10.6\%} & \multicolumn{1}{r}{20.9\%} & \multicolumn{1}{r}{68.6\%} & 0.62 \\  \cline{2-10} 
\multicolumn{1}{l}{\multirow{-3}{*}{All}} & \multicolumn{1}{l|}{$\prompts$} & \multicolumn{1}{r}{12.6\%} & \multicolumn{1}{r}{59.0\%} & \multicolumn{1}{r}{28.5\%} &\multicolumn{1}{r|}{0.33} & \multicolumn{1}{r}{5.5\%} & \multicolumn{1}{r}{48.7\%} & \multicolumn{1}{r}{45.8\%} & 0.41 \\ \hline\hline
 \multicolumn{2}{c|}{Average Consensus}  & \multicolumn{1}{r}{12.8\%}  &\multicolumn{1}{r}{36.7\%}  &\multicolumn{1}{r}{45.1\%} & \multicolumn{1}{c|}{-}  &\multicolumn{1}{r}{10.1\%}  & \multicolumn{1}{r}{30.6\%} & \multicolumn{1}{r}{59.3\%} &
 - \\ 
\end{tabular}}
\vspace{-1em}

\end{table*}

\section{Results and Validation}
\label{results}


\subsection{Annotation Results}
We present the results of our annotation experiments on the five-point Likert scale in Figure \ref{fig:annotation} for the three datasets. These results demonstrate that the preference rate between each pair of systems is highly dependent on the dataset (i.e., the composition of input prompts). We note that the $\MSCOCO$ prompts are the shortest and simplest ones for TTI systems to address. In contrast, the $\prompts$ dataset and the $\LNCOCO$ prompts are longer and more challenging, as shown in previous work \cite{yu2022scaling,saharia2022photorealistic}. 
Observing the first row of Figure \ref{fig:annotation} (\texttt{A} vs \texttt{B}), we note that when prompts are simpler (i.e., $\MSCOCO$), both systems \texttt{A} and \texttt{B} perform well. In fact, in 26\% of the comparisons, annotators rated both systems as ``\texttt{A} and \texttt{B} are the same.'' However, for more challenging datasets, we observed ties amongst 17\% ($\LNCOCO$) and 12\% ($\prompts$) of the comparisons. Consequently, the percentage of extreme choices (i.e., when  one system is preferred ``much'' better than the other) is lowest in $\MSCOCO$, which confirms our previous observation that both systems perform similarly, making it difficult for annotators to distinguish between them. 
On the other hand, in the $\prompts$ dataset, annotators tended to select extreme choices (i.e., one model is "much better" than the other one) more frequently than the other two datasets. Therefore, when the prompts become more challenging, it is easier for annotators to discriminate between two models, as it is more likely that one of the systems fails to satisfy the complex prompt. When the prompts are simple, we hypothesize that both models perform well, and annotators do not perceive many differences between the two sets of generated images.

When comparing the first row with the second row in Figure \ref{fig:annotation} and keeping in mind that \texttt{B'} is the version of \texttt{B} with fewer parameters, we observe that system \texttt{A} was preferred more frequently in the \texttt{A} vs \texttt{B'} comparison rather than in the \texttt{A} vs \texttt{B} comparison across all three datasets. For example, in the $\MSCOCO$ dataset, \texttt{A} was rated as much or somewhat better than \texttt{B} in 43\% of the prompts, while \texttt{A} was preferred over \texttt{B'} in more than 50\% of the prompts. This increase in the preference rate of \texttt{A} from the comparison to \texttt{B} versus the comparison to \texttt{B'}, indicates that, as expected, the larger model \texttt{B} was more competitive with \texttt{A} than the smaller model \texttt{B'}.
It is also interesting that the option "X and Y are the same" received the highest rate when comparing TTI system \texttt{B} with \texttt{B'}. Especially for the $\prompts$ dataset, the two systems are almost indistinguishable and were labeled as the same in over 36\% of the prompts. In comparison, between systems \texttt{B} vs \texttt{B'}, we also note the smallest difference in preference between the two sides, with less than a 4\% difference between when \texttt{B} was preferred over \texttt{B'} and when \texttt{B'} was preferred over \texttt{B}.

\begin{table*}[t!]
\vspace{-1em}
\caption{Agreement rate of the metrics with human annotations. Statistically significant agreement with Wilcoxon paired test and p-value $<$ 0.05 are shown with * symbol. Bold: highest statistically significant agreement in the column.}
\vspace{-1em}
\label{tab:agreement}
\centering
\scalebox{0.9}{
\begin{tabular}{l|rrr|rrr|rrr}
 & \multicolumn{3}{c|}{MSCOCO} & \multicolumn{3}{c|}{LNCOCO} & \multicolumn{3}{c}{prompts} \\ \hline
 & \multicolumn{1}{l|}{\texttt{A} vs \texttt{B}} & \multicolumn{1}{l|}{\texttt{A} vs \texttt{B'}} & \multicolumn{1}{l|}{\texttt{B} vs \texttt{B'}} & \multicolumn{1}{l|}{\texttt{A} vs \texttt{B}} & \multicolumn{1}{l|}{\texttt{A} vs \texttt{B'}} & \multicolumn{1}{l|}{\texttt{B} vs \texttt{B'}} & \multicolumn{1}{l|}{\texttt{A} vs \texttt{B}} & \multicolumn{1}{l|}{\texttt{A} vs \texttt{B'}} & \multicolumn{1}{l}{\texttt{B} vs \texttt{B'}} \\ \hline
Diversity & \multicolumn{1}{r|}{47.2\%} & \multicolumn{1}{r|}{41.9\%} & 53.1\% & \multicolumn{1}{r|}{45.3\%} & \multicolumn{1}{r|}{42.1\%} & 45.7\% & \multicolumn{1}{r|}{45.9\%} & \multicolumn{1}{r|}{46.3\%} & 47.1\% \\ 
rbp & \multicolumn{1}{r|}{53.7\%} & \multicolumn{1}{r|}{55.3\%*} & 55.5\% & \multicolumn{1}{r|}{53.1\%*} & \multicolumn{1}{r|}{64.1\%*} & 52.1\% & \multicolumn{1}{r|}{56.9\%*} & \multicolumn{1}{r|}{60.2\%*} & 54.2\% \\ 
novrbp & \multicolumn{1}{r|}{\textbf{53.8\%}} & \multicolumn{1}{r|}{\textbf{61.2\%*}} & 51.7\% & \multicolumn{1}{r|}{52.5\%*} & \multicolumn{1}{r|}{58.8\%*} & 54.2\% & \multicolumn{1}{r|}{58.5\%*} & \multicolumn{1}{r|}{60.7\%*} & 54.2\% \\ 
urbp & \multicolumn{1}{r|}{53.3\%} & \multicolumn{1}{r|}{58.9\%*} & 52.9\% & \multicolumn{1}{r|}{54.2\%*} & \multicolumn{1}{r|}{60.0\%*} & 53.4\% & \multicolumn{1}{r|}{57.7\%*} & \multicolumn{1}{r|}{60.2\%*} & 49.5\% \\ 
err & \multicolumn{1}{r|}{52.2\%} & \multicolumn{1}{r|}{55.4\%*} & 51.7\% & \multicolumn{1}{r|}{50.0\%*} & \multicolumn{1}{r|}{60.1\%*} & 54.2\% & \multicolumn{1}{r|}{57.7\%*} & \multicolumn{1}{r|}{61.5\%*} & 53.5\% \\ 
noverr & \multicolumn{1}{r|}{52.2\%} & \multicolumn{1}{r|}{60.4\%*} & 50.9\% & \multicolumn{1}{r|}{53.1\%*} & \multicolumn{1}{r|}{60.8\%*} & 56.2\% & \multicolumn{1}{r|}{\textbf{59.3\%*}} & \multicolumn{1}{r|}{\textbf{63.1\%*}} & 54.2\% \\ 
uerr & \multicolumn{1}{r|}{52.2\%} & \multicolumn{1}{r|}{58.4\%*} & 52.9\% & \multicolumn{1}{r|}{\textbf{55.2\%*}} & \multicolumn{1}{r|}{\textbf{62.6\%*}} & 52.7\% & \multicolumn{1}{r|}{58.1\%*} & \multicolumn{1}{r|}{61.5\%*} & 52.3\% \\ 
\end{tabular}}
\end{table*}

 \subsection{Agreement with Human preferences}
Given two sets of images generated by systems $s$ and $s'$, we expect $\evalq > \evalmetric(\system'(\prompt),u_q)$ if human annotators preferred the set of generated images from $s$ over $s'$.
To ensure the reliability of agreement between annotators, we only report the agreement rates of the evaluation metrics on prompts where all three annotators fully agreed on the annotations. 
The agreement rates in Table \ref{tab:agreement} indicate the percentage of the prompts where annotators rate system X as better than system Y \textit{and} our metric also computes $\evalmetric(X(\prompt),u_q) > \evalmetric(Y(\prompt),u_q)$. 
Since FID is not capable of quantifying the quality of a single set of generated images for a given prompt, it is impossible to report the agreement rate between annotators on a per-sample basis.
We also report the results of Wilcoxon paired statistical significance tests with p-values < 0.05 between the annotations and the measured performance by $\evalmetric$.
As shown in Table \ref{tab:agreement}, all six of our proposed metrics are able to show statistically significant agreement rates with annotations when comparing \texttt{A} vs \texttt{B'} on all three datasets. However, they do not show statistically significant agreement with annotators when comparing \texttt{B} vs \texttt{B'}. We hypothesize that since \texttt{B'} is a version of \texttt{B} only varying by the number of parameters, their performance may indeed be quite similar. Therefore, the non-significant agreement rate in this comparison could be the result of actual comparable performance between the two systems. The third row of Figure \ref{fig:annotation} confirms this since the two systems have the highest \textit{"\texttt{B} and \texttt{B'} are the same"} rate (the gray bars) across all three datasets. 

We additionally note that on the two most challenging datasets ($\LNCOCO$ and $\prompts$), we see statistically significant agreement rates with annotators when comparing TTI model \texttt{A} vs \texttt{B}. We note that
\begin{inlinelist}
    \item $\MSCOCO$ prompts are the most simple prompts among the three datasets, and
    \item the $\MSCOCO$ dataset have been widely used for training TTI systems, 
\end{inlinelist}
and so, TTI models are more familiar with the structure of simple prompts and images. However, for more complex prompts, it is more likely that one of the models fails to generate a relevant set of images, making it easier to assess their performance. As shown in Table \ref{tab:agreement}, our proposed metric $\noverr$ outperforms the other evaluation metrics on the $\prompts$ dataset. We achieved a 59.3\% agreement rate with annotators when comparing \texttt{A} vs \texttt{B} and a 63.1\% agreement rate when comparing \texttt{A} vs \texttt{B'}. This metric also shows favorable agreement rates on the other two datasets. Additionally, the position-based metric $\novrbp$, which considers variety and novelty, shows the highest agreement on the $\MSCOCO$ dataset. In general, novelty-based metrics were able to outperform other metrics on all three datasets, confirming the impact of considering variety and novelty in TTI evaluation metrics.

\begin{table*}[]
\label{tabPNFID}
\caption{Distinguishing images from the Preferred ($P$) and Not-preferred ($N$) set of grids in the preference-based comparisons. The lower FID in each comparison indicates the better set. For $\err$ and $\rbp$, higher value indicates better predicted performance. For the sake of easier comparisons, we multiply the output of our metrics by 100. Bold: The better predicted.  }
\vspace{-1em}

\centering

\scalebox{0.9}
{
\begin{tabular}{l|l|rr|rr|rr}
 &  & \multicolumn{2}{c|}{A vs B} & \multicolumn{2}{c|}{A vs B'} & \multicolumn{2}{c}{B vs B'} \\ \cline{3-8} 
\multirow{-2}{*}{Dataset} & \multirow{-2}{*}{Metric} & \multicolumn{1}{l|}{Preferred (P)} & \multicolumn{1}{l|}{Not-preferred (N)} & \multicolumn{1}{l|}{Preferred (P)} & \multicolumn{1}{l|}{Not-preferred (N)} & \multicolumn{1}{l|}{Preferred (P)} & \multicolumn{1}{l}{Not-preferred (N)} \\ \hline
\multicolumn{1}{c|}{} & \cellcolor[HTML]{DADADA}FID & \multicolumn{1}{r|}{\cellcolor[HTML]{DADADA}\textbf{27802.9}} & \cellcolor[HTML]{DADADA}28991.4 & \multicolumn{1}{r|}{\cellcolor[HTML]{DADADA}\textbf{23647}} & \cellcolor[HTML]{DADADA}26586.6 & \multicolumn{1}{r|}{\cellcolor[HTML]{DADADA}\textbf{26673.1}} & \cellcolor[HTML]{DADADA}28757.2 \\
\multicolumn{1}{c|}{} & $\err$ & \multicolumn{1}{r|}{\textbf{98.75}} & 98.69 & \multicolumn{1}{r|}{\textbf{98.79}} & 98.75 & \multicolumn{1}{r|}{\textbf{98.72}} & 98.71 \\
\multicolumn{1}{c|}{} & $\noverr$ & \multicolumn{1}{r|}{\textbf{91.23}} & 90.99 & \multicolumn{1}{r|}{\textbf{91.54}} & 91.35 & \multicolumn{1}{r|}{\textbf{91.35}} & 91.25 \\
\multicolumn{1}{c|}{} & $\rbp$ & \multicolumn{1}{r|}{\textbf{89.01}} & 88.60 & \multicolumn{1}{r|}{\textbf{89.36}} & 88.96 & \multicolumn{1}{r|}{\textbf{88.77}} & 88.69 \\
\multicolumn{1}{c|}{\multirow{-5}{*}{$\MSCOCO$}} & $\novrbp$ & \multicolumn{1}{r|}{\textbf{80.73}} & 80.40 & \multicolumn{1}{r|}{\textbf{81.04}} & 80.75 & \multicolumn{1}{r|}{\textbf{80.64}} & 80.54 \\ \hline
 & \cellcolor[HTML]{DADADA}FID & \multicolumn{1}{r|}{\cellcolor[HTML]{DADADA}24191.6} & \cellcolor[HTML]{DADADA}\textbf{21983.9} & \multicolumn{1}{r|}{\cellcolor[HTML]{DADADA}22671.9} & \cellcolor[HTML]{DADADA}\textbf{21237.3} & \multicolumn{1}{r|}{\cellcolor[HTML]{DADADA}\textbf{23647.0}} & \cellcolor[HTML]{DADADA}26586.6 \\
 & $\err$ & \multicolumn{1}{r|}{\textbf{96.56}} & 96.34 & \multicolumn{1}{r|}{\textbf{96.85}} & 96.38 & \multicolumn{1}{r|}{\textbf{96.23}} & 96.08 \\
 & $\noverr$ & \multicolumn{1}{r|}{\textbf{90.8}} & 90.27 & \multicolumn{1}{r|}{\textbf{91.43}} & 90.35 & \multicolumn{1}{r|}{\textbf{90.08}} & 89.76 \\
 & $\rbp$ & \multicolumn{1}{r|}{\textbf{89.58}} & 89.05 & \multicolumn{1}{r|}{\textbf{90.39}} & 89.04 & \multicolumn{1}{r|}{\textbf{88.66}} & 88.35 \\
\multirow{-5}{*}{$\LNCOCO$} & $\novrbp$ & \multicolumn{1}{r|}{\textbf{64.65}} & 64.27 & \multicolumn{1}{r|}{\textbf{65.04}} & 64.29 & \multicolumn{1}{r|}{\textbf{64.2}} & 63.96 \\ \hline
 & \cellcolor[HTML]{DADADA}FID & \multicolumn{1}{r|}{\cellcolor[HTML]{DADADA}\textbf{3451.5}} & \cellcolor[HTML]{DADADA}4748.6 & \multicolumn{1}{r|}{\cellcolor[HTML]{DADADA}\textbf{3372.1}} & \cellcolor[HTML]{DADADA}4792.3 & \multicolumn{1}{r|}{\cellcolor[HTML]{DADADA}\textbf{9012.0}} & \cellcolor[HTML]{DADADA}9133.8 \\
 & $\err$ & \multicolumn{1}{r|}{\textbf{96.55}} & 96.33 & \multicolumn{1}{r|}{\textbf{96.48}} & 96.29 & \multicolumn{1}{r|}{\textbf{96.43}} & 96.38 \\
 & $\noverr$ & \multicolumn{1}{r|}{\textbf{90.91}} & 90.41 & \multicolumn{1}{r|}{\textbf{90.76}} & 90.27 & \multicolumn{1}{r|}{\textbf{90.55}} & 90.49 \\
 & $\rbp$ & \multicolumn{1}{r|}{\textbf{89.47}} & 88.86 & \multicolumn{1}{r|}{\textbf{89.29}} & 88.76 & \multicolumn{1}{r|}{\textbf{89.1}} & 88.99 \\
\multirow{-5}{*}{$\prompts$} & $\novrbp$ & \multicolumn{1}{r|}{\textbf{64.85}} & 64.5 & \multicolumn{1}{r|}{\textbf{64.73}} & 64.39 & \multicolumn{1}{r|}{\textbf{64.63}} & 64.61 
\end{tabular}
\vspace{-1em}
\label{tabPNFID}
}
\vspace{-1em}
\label{tabPNFID}
\end{table*}

\subsection{Ablation Study} We  studied the impact of $\prob(\trajectory)$ (Section \ref{trajectory}) on the agreement rates by comparing $\novrbp$ with $\urbp$ and similarly comparing $\noverr$ with $\uerr$. We observe in Table \ref{tab:agreement} that $\noverr$ shows a superior agreement rate compared to $\uerr$ on the majority of datasets when the agreement rate is statistically significant. Additionally, $\novrbp$ also shows a greater agreement rate compared to $\urbp$ on two out of the three datasets. By comparing the impact of $\prob(\trajectory)$ across the three datasets, we hypothesize that sampling $\trajectory$ from the saliency distribution of the grid works best as the prompt sets become more challenging and it has the most positive effect on the $\prompts$ dataset and less positive effect on the easier prompt sets.
We have also included the agreement rate of the $Diversity$ metric. However, it failed to show any statistically significant agreement with the human annotators. Nonetheless, by comparing the performance of metrics with and without the variety and novelty components, we can conclude that incorporating variety and novelty in TTI evaluation metrics improves the agreement rates of both the position-based and cascade-based metrics with the human annotations \cite{arabzadeh2023adele}.


\vspace{-1em}
\subsection{Comparison with FID}
Although we cannot measure prompt-level agreement for FID, we can measure the population-level agreement.  If comparing two systems $X$ and $Y$, we can use the set of all images in the Preferred grids \textit{regardless of the source system}; call this set ($P$). 
Similarly, we have the set of images in Non-preferred grids  ($N$). We measure FID between the following sets: 1) set of target images and all the generated images from the Preferred grids ($P$) and 2) set of target images and all the generated images in Non-preferred grids ($N$). 
We expect that the set of Non-preferred grids to show higher FID (lower similarity) with the target images. 
Our results (Table \ref{tabPNFID}) show that, while FID is able to measure better performance (lower distance with target images) on the simplest set of prompts (i.e., $\MSCOCO$), it fails to detect the better-performed set for  challenging prompt sets. In fact, for $\LNCOCO$, FID indicated that the Non-preferred sets of generated images had performed better when comparing both \texttt{A} vs \texttt{B} and \texttt{A} vs \texttt{B'}. However, it correctly measured a lower FID score for the preferred set in the comparison of \texttt{B} and \texttt{B'}. 

We can compare the population-level consistency of FID with that of our proposed metrics. 
Similar to the FID analysis, we report the average of our metrics on both P and N sets of generated images. Unlike FID, all of our metrics were able to assign higher values when comparing any pairs of the TTI systems and on all the three datasets. All four of our metrics consider the generated images from Preferred grids (P) as the better set compared to the images from Non-preferred grids (N).

\vspace{-1em}
\section{Conclusion}
\label{conclusion}
Because reliable offline evaluation metrics for TTI systems is critical for benchmarking purposes in the community \cite{otani:tti-evaluation-status}, we proposed a set of evaluation metrics for set-based TTI inspired by grounding ideation criteria in traditional IR evaluation methods.  
We use existing and novel approaches to capture fluency, variety, novelty, and quality, while incorporating visual salience as a fundamental feature of image layouts.  Moreover, to the best of our knowledge, this is the first work to explore assessing generated images as a \textit{set}.  We demonstrated the effectiveness of our family of metrics against the popular FID metric, showing that ideation-based measures were better aligned with human preferences.  
Although our results are positive, we believe there are many directions to continue offline TTI evaluation work.  First, our models of visual salience and browsing, while validated on human preferences, could benefit from models customized for ideation tasks.  Second, ideation and generation tends to be highly interactive, consisting of multiple turns where state can be carried across individual sessions and even between users.  This complex interaction suggests that single turn metrics provide an incomplete picture of system effectiveness.  Finally, our results may provide insight into how to design \textit{online} evaluation metrics for TTI based on more elaborate behavior and interaction than found in classic information retrieval settings.
\bibliographystyle{ACM-Reference-Format}
\balance
\bibliography{acmart} 

\end{document}